\documentclass[conference,anonymous,review]{IEEEtran}
\IEEEoverridecommandlockouts
\usepackage{cite}
\usepackage{amsmath,amssymb,amsfonts}
\usepackage{algorithm}
\usepackage{algpseudocode}
\usepackage{graphicx}
\usepackage{textcomp}
\usepackage{cite}
\usepackage{xcolor}
\usepackage{svg}
\usepackage{tikz}
\usepackage[normalem]{ulem}
\usepackage{pgfplots}
\usepackage{hyperref}
\usepackage{array}
\usepackage{subcaption}
\usepackage{multirow}

\newcolumntype{P}[1]{>{\centering\arraybackslash}p{#1}}
\newcolumntype{M}[1]{>{\centering\arraybackslash}m{#1}}

\usetikzlibrary{positioning}
\def\BibTeX{{\rm B\kern-.05em{\sc i\kern-.025em b}\kern-.08em
    T\kern-.1667em\lower.7ex\hbox{E}\kern-.125emX}}

\newif\ifannotated
\annotatedtrue %

\usepackage{resources/mathcommands}

\usepackage{resources/scheduling}

\newtheorem{theorem}{Theorem}

\begin{document}

\title{Bridging the Gap between ROS~2 and Classical Real-Time
  Scheduling for Periodic Tasks}

\author{
  Harun Teper$^1$, Oren Bell$^2$, Mario Günzel$^1$, Chris Gill$^2$,
  and Jian-Jia Chen$^1$$^3$\\
  $^1$ TU Dortmund University, Germany\\
  $^2$ Washington University in St Louis, USA\\
  $^3$ Lamarr Institute for Machine Learning and Artificial
  Intelligence, Germany\\
  Emails: \{harun.teper, mario.guenzel, jian-jia.chen\}@tu-dortmund.de
  and \{oren.bell, cdgill\}@wustl.edu
}

\maketitle

\begin{abstract}
The Robot Operating System 2 (ROS~2) is a widely used middleware that provides software libraries and tools for developing robotic systems.
In these systems, tasks are scheduled by ROS~2 executors.
Since the scheduling behavior of the default ROS~2 executor is inherently different from classical real-time scheduling theory, dedicated analyses or alternative executors, requiring substantial changes to ROS~2, have been required.

In 2023, the events executor, which features an \emph{events queue} and allows the possibility to make scheduling decisions immediately after a job completes, was introduced into ROS~2. 
In this paper, we show that, with only minor modifications of the events executor, a large body of research results from classical real-time scheduling theory becomes applicable. 
Hence, this enables analytical bounds on the worst-case response time and the end-to-end latency, outperforming bounds for the default ROS 2 executor in many scenarios.
Our solution is easy to integrate into existing ROS 2 systems since it requires only minor backend modifications of the events executor, which is natively included in ROS 2.
The evaluation results show that our ROS~2 events executor with minor modifications can have significant
improvement in terms of dropped jobs, worst-case response time, end-to-end latency, and performance compared to the default ROS~2 executor.
\end{abstract}

\section{Introduction}\label{sec:intro}

The Robot Operating System 2 (ROS~2) is a widely used middleware for creating robotics applications.
It provides tools and libraries to build modular systems consisting of many interacting components.
In comparison to the original Robot Operating System (ROS), it provides opportunities to configure real-time properties through its use of the Data Distribution Services (DDS) for real-time communication and its custom scheduling abstraction, called an executor, that manages the execution of time-triggered and event-triggered tasks.

The standard scheduling mechanism in ROS 2 relies on \emph{polling points} and \emph{processing windows}. 
That is, at each polling point, jobs that are eligible for execution are moved to a \emph{wait set}, followed by a processing window within which to schedule the jobs in the wait set non-preemptively by a predefined priority order. 
Due to this round-robin-like approach, even tasks with a high priority may experience long blocking times.
That is, a high-priority task may be blocked for the length of a \emph{full} processing window --- including the execution time of \emph{every} lower-priority task. 
Moreover, the implementation of ROS 2 only releases one job of a task, even if the task period has been reached several times during one processing window.

Dedicated analyses for the ROS 2 standard executor have been provided~\cite{casini2019response,tang2020response,blass2021,teper2022,teper2023}.
To mitigate potentially long response times of the ROS~2 default executor,
alternative executor designs have been proposed and analyzed, such as fixed-priority schedulers~\cite{choi2021picas} and dynamic-priority schedulers~\cite{arafat2022}.
However, they can not be integrated easily into existing ROS 2 systems, as they require substantial changes in the source code of ROS 2.

In 2023, the events executor~\cite{eventexecutor} was introduced in ROS~2, replacing the wait set with an \emph{events queue}, and passing only one job to the processing window at a time. 
This paper addresses the following fundamental question:
\begin{quote}
  \emph{Is it possible to use the events executor in ROS~2 in a manner that is compatible with the classical real-time scheduling theory of periodic task systems, in which every high-priority task can only be blocked by at most \emph{one} lower-priority task, and a job is released \emph{every time} the task period is reached?}
\end{quote}
Our answer is \emph{yes, but only under certain conditions.} 

In this paper, we uncover these conditions and show that any priority-based non-preemptive scheduling strategy, with slight modifications of the events executor, can be realized, including Fixed-Priority (FP) and Earliest-Deadline-First (EDF) scheduling.
Our solution is easy to integrate into existing ROS~2 systems since it requires only minor backend modifications of the events executor, which is natively included in ROS~2.
With our solution, typical results from the real-time systems literature for priority-based non-preemptive scheduling of periodic tasks can be applied.

\noindent\textbf{Our Contributions}: 
The contributions of this paper are:
\begin{itemize}
\item In Section~\ref{sec:background}, we outline the different characteristics
  of typical priority-based schedulers and the default ROS~2 executor. 
  We define the problem this paper investigates in Section~\ref{sec:problem-definition}.
  The events executor is introduced in Section~\ref{sec:events_executor}.
\item We present modifications to allow priority-based scheduling for the events executor in Section~\ref{sec:modifications}.
\item We state the conditions that make our proposed executor compatible with priority-based non-preemptive scheduling theory for periodic tasks in Section~\ref{sec:compatibility}. Furthermore, we demonstrate how the analytical results in the literature can be applied to bound typical metrics like the worst-case response time and the end-to-end latency for the proposed ROS 2 scheduler.
\item While the previous sections focus on timer tasks (i.e., tasks with periods), we extend our paper to subscription tasks (i.e., tasks triggered by other tasks) in Section~\ref{sec:subscription}.
\item In Section~\ref{sec:experiments} we evaluate our findings, showing our compatibility with the classical real-time scheduling theory and the benefits of our proposed scheduler. We show that response-time bounds from the literature can be directly applied to ROS~2 systems, improving end-to-end latencies significantly in many cases.
\end{itemize}

\section{Periodic Task Systems and Executors in ROS~2}\label{sec:background}

In this section, we first present the basic model widely used in
real-time scheduling theory for periodic task systems,
followed by the default ROS~2 executor.
For simplicity
of presentation, we focus on ROS~2 timers in this section.

\subsection{Typical Priority-Based Scheduler}\label{sec:periodic-taskmodel}

We first introduce the widely adopted periodic task model in
real-time systems. Given a set $\Tbb$ of tasks, a
\textbf{periodic task} $\tau_i$ is specified by the tuple
$\tau_i = (C_i, T_i, D_i, \phi_i) \in \Rbb^4$, where $C_i\geq 0$ is
the worst-case execution time (WCET), $T_i>0$ is the period, $D_i > 0$
is the relative deadline, and $\phi_i$ is the phase. The periodic
task $\tau_i$ releases its first job (task instance) at time~$\phi_i$, and
subsequent jobs are released every $T_i$ time units.  Every job of $\tau_i$
executes for at most $C_i$ time units and has an absolute deadline
specified as its release time plus the relative deadline. 
When a job is released at time $t$, it is placed into the ready queue and the job
is removed from the ready queue after it finishes executing.
We restrict our paper to \emph{constrained deadline systems}, where
$D_i \leq T_i$ for all tasks $\tau_i\in \Tbb$.

In real-time operating systems (RTOSes), the release of the jobs of
the periodic tasks can be implemented as a job \emph{releaser}\footnote{We use the term \emph{releaser} to indicate an independent component of a system that is responsible for determining whether and if so when a job should be released.  Its role is distinct from that of the \emph{scheduler}, which decides which of the released jobs should be running at any point in time.} based on a
\emph{timer interrupt service routine}. Specifically, suppose that
$T_i$ is an integer multiple of a predefined system tick for every task
$\tau_i\in \Tbb$. A timer interrupt is triggered every tick and the
releaser has to decide whether a job of task $\tau_i\in \Tbb$ should
be released or not.

A \emph{scheduler} determines which job in the ready queue should be
executed. \emph{Priority-based schedulers}, in which scheduling
decisions are made by assigning priorities to jobs, have been widely
studied in the literature.  At any point in time when a scheduling
decision has to be made, the \emph{highest-priority job among the jobs
  in the ready queue} is allocated to the processor for execution. In
the literature, such priority-based schedulers (on the task level) are
classified into fixed-priority and dynamic-priority schedulers. A
scheduler is a \emph{fixed-priority scheduler} if, for any two tasks
$\tau_i$ and $\tau_j$, either all jobs of $\tau_i$ have higher
priority than all jobs of $\tau_j$ or all jobs of $\tau_j$ have higher
priority all jobs of $\tau_i$. In contrast, for
\emph{dynamic-priority} scheduling, a job of $\tau_i$ may have a
higher priority than some jobs of $\tau_j$ but a lower priority than other jobs of $\tau_j$.

Specifically, the rate monotonic (RM) scheduler (where a task with a
smaller period has a higher priority) is a well-known fixed-priority
scheduling policy, whilst the earliest-deadline-first (EDF) scheduler
(where a job with the earliest absolute deadline has the highest priority)
is a well-known dynamic-priority scheduling policy~\cite{liu73scheduling}.

Scheduling algorithms also may be preemptive or non-preemptive.
That is, while for preemptive scheduling a scheduling decision is made at every timer interrupt, for non-preemptive scheduling decisions are only made at a few specific checkpoints of job execution.

More specifically, in preemptive scheduling, at every timer interrupt it is checked whether the currently running job is still the highest-priority job in the ready queue, and if so, it continues executing.
Otherwise, the system performs a context switch, preempts the currently executing job, and executes the new highest-priority job in the ready queue instead.

On the other hand, for non-preemptive scheduling, a scheduling decision is only made when (i) a running job finishes its execution or (ii) a job is inserted into an empty ready queue.
Since the scheduler can make a scheduling decision to start the execution of the highest-priority available job only at those time points, a running job is never preempted and continues to be executed until it finishes.
Worst-case analysis of non-preemptive schedulers for
real-time systems has been widely studied, e.g.,
\cite{DBLP:journals/rts/DavisTGDPC18,DBLP:journals/rts/DavisBBL07,george1996EDFNP,DBLP:conf/rtss/NasriB17,DBLP:conf/rtss/JeffaySM91}.

\subsection{Default Executor in ROS~2}\label{sec:defaultexecutor}

Next, we describe the default executor in ROS~2, and how it relates to the periodic task model presented in Section~\ref{sec:periodic-taskmodel}.

The default ROS~2 executor non-preemptively schedules tasks' eligible jobs using a wait set.
Its scheduling mechanism is split up into two iterative phases, the \emph{polling point} and the \emph{processing window}.
At each polling point, the executor updates the wait set by sampling at most one job of each eligible task.
During each processing window, the executor iterates over the wait set and executes the jobs non-preemptively according to the fixed-priority order of their respective tasks.
The executor thread executes a job by calling a function (callback).
After finishing the execution of all jobs in the wait set, the wait set is empty, and the executor starts a new polling point.

Compared to classical real-time schedulers, the polling point describes the moment when tasks' jobs are released, and the processing window describes the interval over which the released jobs are scheduled.

Furthermore, timers are handled differently in ROS~2 compared to the classical real-time releaser.
Instead of actively being inserted into the wait set by a releaser at the time of release, timers are passively sampled by the executor at the polling point by determining if they are eligible for execution.

The mechanism of the default executor for timers is as follows:
Each timer stores its next \emph{activation time} at which it will be eligible for execution.
At the moment at which the executor determines if a timer is eligible, it checks whether the current time is greater than or equal to the \emph{activation time} of the timer. If so, a job of that timer is sampled, and the next \emph{activation time} point is determined by increasing the current \emph{activation time} by the minimal multiple of the timer period, such that it is greater than the current time.
As a result, the executor may skip timer jobs if the time between two polling points is longer than the period of the timer, as the executor may increase the \emph{activation time} of the timer by a multiple of the period.
For the rest of the paper, we refer to the \emph{activation time} as the \emph{timestamp} of the timer job.

In general, this combination of a non-preemptive scheduler using a wait set and the passive sampling of timers can lead to interference between the release and scheduling of a job.
For the default executor, the \emph{timestamp} is incremented at the time of scheduling the timer job, instead of the time at which the job is sampled and added to the wait set.
Potentially, this leads to the timer being prevented from releasing jobs at its specified period, if the processing window is long enough to prevent the executor from updating the \emph{timestamp} of the timer before the next period elapses.

While this design may have disadvantages in guaranteeing the periodic release of timers, it ensures that all jobs that are polled into the wait set are executed in the subsequent processing window.
That is, a timer may not generate as many jobs as its period indicates, but whenever a job is ready it is executed in the following processing window.
Although this may lead to tasks being executed less often than expected, it improves ease of development and also makes it possible to set a very short timer to ensure that some actions are performed as frequently as possible, e.g., by setting a timer to $0$.

As the default ROS~2 executor (based on the concept of polling points and processing windows in an interleaved manner) does not match the model (based on an independent \emph{releaser} and \emph{scheduler}) of the classic real-time scheduling theory presented in Section~\ref{sec:periodic-taskmodel}, the rich literature of real-time scheduling theory is not applicable for ROS~2 systems that use the default executor.
Hence, dedicated response-time analyses have been developed for the default ROS~2 executor~\cite{casini2019response,tang2020response,blass2021}, and suggestions have been made on how to optimize the ROS~2 application code to mitigate long response times~\cite{tang2020response}.
Casini et al.~\cite{casini2019response} also introduced the concept of processing chains, in which tasks may be triggered due to the arrival of data, and output a result, triggering other tasks.
This propagation of data creates a natural structure of chained tasks, that can be used to measure end-to-end latency of the task set.
Specifically, Teper et al. have analyzed~\cite{teper2022,teper2024rtas} and optimized~\cite{teper2023} such end-to-end latencies in the ROS~2 single-threaded executor.
Multi-threaded executors have been developed and analyzed~\cite{jiang2022,sobhani2023,lange2018cbgexecutor}, but these are still constrained by the wait set behavior of the default executor. Their implementation either shares the wait set among threads in a thread pool or runs multiple unmodified executors in different threads. The latter approach also requires the manual assignment of nodes to executors.

Although incompatibility with classical real-time schedulers can be addressed by designing alternative executors for fixed-priority~\cite{choi2021picas} or dynamic-priority~\cite{arafat2022} schedulers as we have noted earlier, the ability to use executors provided directly by ROS 2 for periodic task systems is still desirable. It reduces design, implementation, and maintenance costs for system developers. Section~\ref{sec:events_executor} describes the new ROS 2 events executor, which offers just such an opportunity.

\section{Problem Definition}\label{sec:problem-definition}

As we have discussed in Section~\ref{sec:defaultexecutor}, the default executor of ROS~2 is not compatible with typical literature results on real-time systems. 
That is, the default executor shows a behavior that is typically not considered when looking at typical priority-based schedulers.
More specifically, typical priority-based schedulers have the following properties:
\begin{itemize}
  \item Each job is inserted into the ready queue (almost) immediately by the releaser.
  \item Each scheduling decision determines only one job to be executed next.
\end{itemize}
Contrarily, the default scheduler of ROS~2 works as follows:
\begin{itemize}
  \item A job is inserted into a wait set only at polling points, and only if the corresponding task has been activated since the preceding polling point.
  \item At the polling point, the scheduler makes the scheduling decision to run all jobs in the wait set in a deterministic order and does not make another scheduling decision until the next polling point.
\end{itemize}
As a direct consequence, the default ROS~2 scheduler shows behavior that usually can not occur in typical priority-based schedulers. 
For example, a job release can be delayed for a whole processing window, including the execution of all lower-priority jobs, and
job releases can even be skipped if a processing window takes longer than a task period.

In this paper, we study the possibility of utilizing the events executor of ROS~2, introduced in Section~\ref{sec:events_executor}, to link to the classical real-time scheduling theory of periodic task systems.
We discuss the necessary modifications of the events executor in Section~\ref{sec:modifications} and discuss the compatibility of the proposed executor in Section~\ref{sec:compatibility}.
While Sections~\ref{sec:modifications} and~\ref{sec:compatibility} focus on applications that are composed of \emph{only timer tasks}, we extend our results to ROS~2 applications with non-timer tasks (i.e., callbacks based on the publisher/subscriber model) in Section~\ref{sec:subscription}.

\section{The Events Executor}
\label{sec:events_executor}

In 2023, ROS~2 added the events executor to its distribution~\cite{eventexecutor}, providing an alternative to the default executor.
It does not use a wait set, but instead uses a FIFO queue, called an \emph{events queue}, that stores the jobs of tasks that are eligible for execution.
It uses two threads, the \emph{timer management thread} and the \emph{default thread}, to manage the release and scheduling of tasks jobs, respectively.
Non-timer tasks are by default only handled by the \emph{default thread}, including releasing and scheduling them.
For timer tasks, the \emph{timer management thread} manages a \emph{timer release queue}, and there are two options for how it handles the tasks in the queue:
\begin{itemize}
  \item Option 1 --- \textbf{Release-Only} (denoted as \textbf{RO}):  The \emph{timer management thread} only releases the timer jobs from the \emph{timer release queue} to the \emph{events queue} (where also non-timer events reside), and all jobs are scheduled by the \emph{default thread} in FIFO order.
  \item Option 2 --- \textbf{Release-and-Execute} (denoted as \textbf{RE}):
  The \emph{timer management thread} holds both released and unreleased timer jobs in its \emph{timer release queue}, and orders and schedules them based on their \emph{timestamp}.
\end{itemize}

Option \textbf{RO} is more in line with the classical real-time scheduling theory, as it separates the \emph{releaser} and \emph{scheduler} for timers, while Option \textbf{RE} is more in line with the default executor, as it does not separate them.
However, neither option incorporates priority-based ordering.
In the following, we discuss the behavior of the scheduler under both options.

Algorithm~\ref{alg:releaser} summarizes the implementation of the \emph{timer management thread} in ROS~2 using the \emph{timer release queue}.
In Algorithm~\ref{alg:releaser}, the \emph{timers} variable refers to the \emph{timer release queue}, which stores the timers ordered by their \emph{timestamp}.

We now describe the algorithm in detail.
For each release, the \emph{timer management thread} checks the head of the queue in Lines~4-5, as the timers are ordered by their \emph{timestamp}.
If the current time is greater than or equal to the \emph{timestamp} of the timer, the timer is considered eligible.
When a timer is released, the \emph{timestamp} of the timer is updated to the next \emph{timestamp} in Line~6, which is the smallest multiple of the period that is greater than the current time.
For the \textbf{RO} option, the timer job is released into the \emph{events queue} in Line~8.
Likewise, for the \emph{RE} option, the timer job is immediately scheduled in Line~10.
Afterward, in Line~12, as the \emph{timestamp} is updated, the queue is reordered according to the new \emph{timestamps} of the tasks in the queue.

Contrary to the default executor, for the \textbf{RO} option, the release is separated from the thread that schedules the timer jobs.
Furthermore, the events executor updates the \emph{timestamp} of the timers between each release.
This guarantees that all future scheduling decisions are based on the most recent \emph{timestamps} of the timers.

In both options, the \emph{timer management thread} is responsible for releasing the timer jobs. The events executor does not specify the priorities of the \emph{timer management thread} and the \emph{default thread}.

As we intend to ensure that the releaser can release the jobs of a timer periodically, in our design and discussions, we assign
a higher priority to the \emph{timer management thread} to ensure that the timer release can not be delayed by the scheduling of non-timer tasks. We note that this is not a limitation, but a treatment that we propose.

For both \textbf{RO} and \textbf{RE} options, we will highlight the effects of this assumption on the scheduling of the timer jobs.In the following, we discuss the behavior of the scheduler under both options in detail.
For simplicity of presentation, we focus on ROS~2 timers in this section.

\begin{figure}[t]
  \centering
  \includegraphics[width=0.45\textwidth]{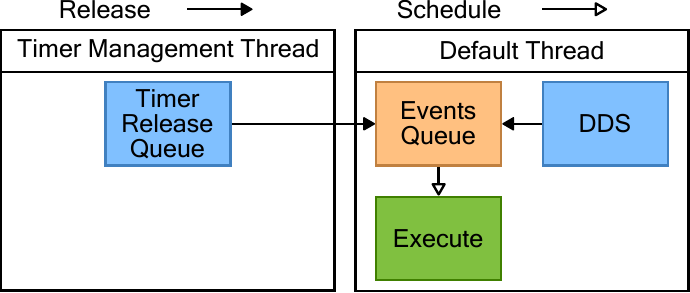}
  \caption{Release-Only model of the events executor}
  \label{fig:release_only}
\end{figure}

\begin{figure}[t]
  \centering
  \includegraphics[width=0.45\textwidth]{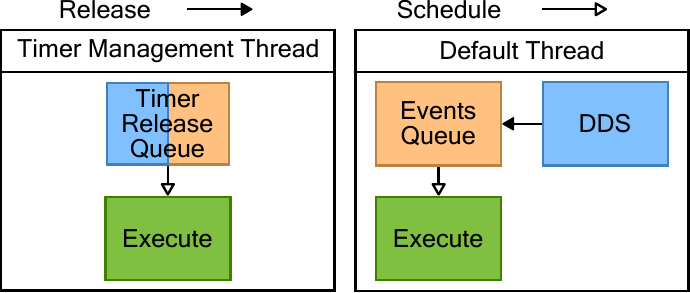}
  \caption{Release-and-Execute model of the events executor}
  \label{fig:release-and-execute}
\end{figure}

\begin{algorithm}[]
  \begin{algorithmic}[1]
    \small 
    \While{Running}
      \State $time\_to\_sleep \gets$ next timer release time
      \State $wait\_for(time\_to\_sleep)$

      \State head\_timer $\gets$ timers.front()

      \While{head\_timer is eligible}
        \State head\_timer.update\_timestamp()

        \If{Option \textbf{RO} is configured}
          \State $events\_queue.enqueue(head\_timer, data)$
        \ElsIf{Option \textbf{RE} is configured}
          \State $execute(head\_timer, data)$
        \EndIf

        \State Reorder timers by release time

        \State head\_timer $\gets$ timers.front()
      \EndWhile
    \EndWhile
  \end{algorithmic}
  \caption{\emph{Timer Management Thread} in Events Executor}
  \label{alg:releaser}
\end{algorithm}

\subsection{Release-Only (\textbf{RO}) Option}\label{sec:release-only}

We first consider the \textbf{RO} option of the events executor, where the \emph{timer management thread} only releases the timer jobs to the \emph{events queue}, while the \emph{default thread} is responsible for scheduling the released timer jobs in the \emph{events queue}.
Figure~\ref{fig:release_only} illustrates the mechanism of the \textbf{RO} option.
The blue color represents a data structure holding unreleased timer jobs, the orange color represents released but not-yet-scheduled timer jobs, and the green color represents scheduled timer jobs.

For this option, the \emph{default thread} can be in one of three states, \emph{waiting}, \emph{suspended}, or \emph{executing}.
Initially, the thread is \emph{waiting}, until the \emph{timer management thread} releases a job into the \emph{events queue}.
Once a job is released, the \emph{default thread} selects the first job from the \emph{events queue} and executes it, transitioning to the \emph{executing} state.
In the \emph{executing} state, the \emph{default thread} iteratively selects the next job from the \emph{events queue} and executes it (non-preemptively), until no more eligible jobs are in the \emph{events queue}.
Once the queue is empty, the \emph{default thread} transitions back to the \emph{waiting} state.

For the scheduling of the threads, we need to consider the priorities of the \emph{timer management thread} and the \emph{default thread}.
Specifically, we assume the \emph{timer management thread} has a higher priority than the \emph{default thread}.
As a result, every release of a timer job by the \emph{timer management thread} can preempt the \emph{default thread} when it is executing a job, leading to the \emph{default thread} transitioning to the \emph{suspended} state.

We note that the above discussion simplifies the interactions between the \emph{default thread} and the \emph{timer management thread} by ignoring the mutual exclusion (mutex) lock that is used to ensure no race conditions when using the \emph{events queue}. Therefore, the \emph{timer management thread} also can be blocked by the \emph{default thread} if the mutex has been locked already by the \emph{default thread}.  Such blocking is considered part of the job release overhead and is ignored for the rest of the paper. 

\subsection{Release-and-Execute (\textbf{RE}) Option}\label{sec:release-and-execute}

Next, we consider the \textbf{RE} option of the events executor, where the \emph{timer management thread} is responsible for both releasing and executing the timer jobs.
The mechanism of this option is illustrated in Figure~\ref{fig:release-and-execute}.

As in the \textbf{RO} option, the \emph{timer management thread} uses the \emph{timer release queue} to store the timers ordered by their \emph{timestamp}.
The \emph{timer release queue} is responsible for storing both the unreleased and released timer jobs.
This is possible, as the timers in the queue are ordered by their \emph{timestamp}, and the \emph{timer management thread} can directly select the first timer in the queue for execution.

Non-timer tasks are released and scheduled by the \emph{default thread} in the \textbf{RE} option, as in the \textbf{RO} option.
Therefore, if a system only has timer tasks, the \emph{default thread} is not used.

For this option, the \emph{timer management thread} can be in one of two states, \emph{waiting}, or \emph{executing}.
When the \emph{timer management thread} is \emph{waiting}, it waits for the release of the next timer, given by the wake-up time of the first timer in the queue.
After a job is released, the \emph{timer management thread} transitions to the \emph{executing} state, where it selects the first timer from the \emph{timer release queue} and executes it.
In this state, timers are executed until the queue does not have any more eligible timers.
After there are no more eligible timers, the \emph{timer management thread} transitions back to the \emph{waiting} state.

As in the \textbf{RO} option, the \emph{timer management thread} updates the \emph{timestamp} of the timer at the moment of releasing the timer job and reorders the \emph{timer release queue} finishing the execution of the timer job.
Thus, each release considers the most recent \emph{timestamps} of the timers, but in contrast to the \textbf{RO} option, the release of timer jobs may be delayed, if the \emph{timer management thread} is executing a timer job.

\section{Modifications to Allow Priority-Based Scheduling}
\label{sec:modifications}

With its current implementation, the events executor executes the timers in FIFO ordering only.
This section presents the modifications that are necessary to execute timers with RM and EDF prioritization. 

Our main proposal is to manage the release of timer jobs and the scheduling of timer jobs using separate queues.
Furthermore, we propose to use a priority queue to store the released timer jobs, which allows for arbitrary fixed-priority or dynamic-priority non-preemptive scheduling policies.

\subsection{Release-Only (\textbf{RO}) Option}

For the  \textbf{RO} option, there are already different queues for managing the release and scheduling of timers. 
That is, the \emph{timer management thread} manages the release of timer jobs separately from the \emph{default thread}, and all released jobs are stored in the \emph{events queue}.
Furthermore, the insert operation of the priority-based \emph{events queue} does not simply insert the job at the end of the queue but inserts it at the correct position according to the priority of the job, which allows for arbitrary fixed-priority or dynamic-priority scheduling policies.

To ensure that timers are moved from the release queue to the \emph{events queue} (almost) immediately, we assign a higher priority to the \emph{timer management thread}, so that the job releases are not blocked by the \emph{default thread}.
As a result, there may be suspensions of the \emph{default thread}, but the timer jobs can be released periodically, as detailed in Section~\ref{sec:compatibility}.

\subsection{Release-and-Execute (\textbf{RE}) Option}

\begin{figure}[t]
  \centering
  \includegraphics[width=0.45\textwidth]{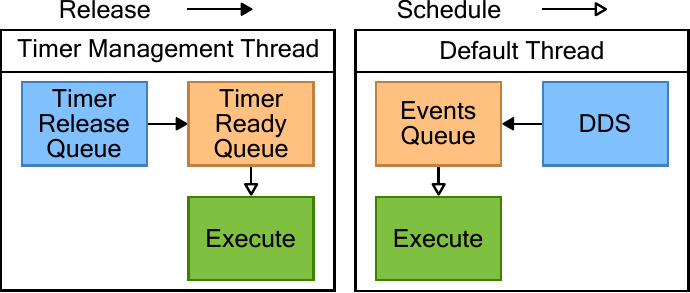}
  \caption{Prioritized Release-and-Execute model of the events executor}
  \label{fig:priority-scheduling}
\end{figure}

For the \textbf{RE} option, we propose to add a priority-based queue to the \emph{timer management thread}, which stores the timer jobs that have been released for execution, i.e., a prioritized ready queue.
We refer to this queue as the \emph{timer ready queue}.
The new design for the \textbf{RE} option is illustrated in Figure~\ref{fig:priority-scheduling}.
Unreleased timer jobs are stored in the \emph{timer release queue}, indicated by the blue color, and released timer jobs in the \emph{timer ready queue}, represented by the orange color.

Furthermore, all eligible timer jobs are inserted into the \emph{timer ready queue} in a priority-based manner.
Therefore, all scheduling decisions consider the priority of all released timer jobs, and the highest-priority job is selected for execution.
The \emph{timer release queue} is reordered after each release, while the \emph{timer ready queue} is ordered through priority-based insertions.

\subsection{Priority-based Scheduling}\label{sec:priority-scheduling}

We now discuss how to enable priority-based scheduling for timer tasks in ROS~2 using the events executor.
For both options, the proposed design is compatible with the classical real-time scheduling theory.
Specifically, the releasing and scheduling of jobs are separated, and the executor makes scheduling decisions between the execution of timer jobs.
Furthermore, the executor uses a  \emph{priority-based events queue} to prioritize the timer jobs.

Our method only adds an overhead of $O(\log n)$ for each priority insert, where $n$ is the number of tasks in the system. This overhead is necessary for priority-based management, unless special treatments/assumptions are made for priority-based management. It is negligible compared to other ROS~2 management routines.

The proposed design is theoretically compatible with any job-level fixed-priority non-preemptive scheduler if the priority can be decided when the job arrives, e.g., the non-preemptive rate-monotonic (RM) and earliest-deadline-first (EDF) scheduling policies for uniprocessor systems.
For RM, the executor can use the timer period as the priority, and for EDF with implicit deadlines, the executor can use the next \emph{timestamp} of the timer as the priority. We note that our current implementations support any task-level fixed-priority non-preemptive scheduler as well as EDF (for implicit deadlines) which are fully compatible with the current ROS~2 codebase.

\section{Compatibility with Non-Preemptive Schedulers for Periodic Tasks}\label{sec:compatibility}

In this section, we explain how our proposed scheduler from Section~\ref{sec:modifications} bridges the gap between the literature on priority-based scheduling and ROS~2.
We assume that the ROS~2 application has only timers and the system is exclusively used to execute the ROS~2 application on one processor.
Furthermore, we assume that tasks have different priorities and ties are broken arbitrarily but deterministically.
We show that our scheduler behaves analytically \emph{like} a non-preemptive work-conserving priority-based scheduler in Section~\ref{sec:compatibility:behavior}.
In Section~\ref{sec:compatibility:overhead} we discuss the overhead that results from our scheduler.
We derive a worst-case response time analysis in Section~\ref{sec:compatibility:wcrt} and 
an end-to-end latency bound (which is one of the predominantly studied metrics for ROS~2 systems) in Section~\ref{sec:compatibility:e2e}, using existing literature results on priority-based real-time scheduling.

\subsection{Analytical Behavior of Our Scheduler}
\label{sec:compatibility:behavior}

Our solution uses the events executor with a modification of the \emph{events queue} for priority-based job ordering.
In the following, we examine both options, Release-Only (\textbf{RO}) and Release-and-Execute (\textbf{RE}), individually and explain how they mimic typical priority-based schedulers.

\paragraph{Release-Only (\textbf{RO})}

With the \textbf{RO} option, the release of timer jobs into the \emph{events queue} is done in a separate thread. 
Therefore, under the assumption that the \emph{timer management thread} is not blocked whenever the \emph{timestamp} of the timer task is reached, a job of that task is inserted into the \emph{events queue} immediately.
Therefore, with \textbf{RO}, jobs are released (i.e., inserted into the \emph{events queue}) periodically.
Moreover, scheduling decisions are effectively made after a job finishes, by pulling the first job (i.e., the highest priority job) from the \emph{events queue}.
Therefore, our scheduler with \textbf{RO} behaves like a typical non-preemptive scheduler.

\paragraph{Release-and-Execute (\textbf{RE})}

With the \textbf{RE} option, the \emph{timer management thread} manages the release of timer jobs using the \emph{timer release queue}, and the scheduling of timer jobs with the \emph{timer ready queue}.
Therefore, the execution of a task can block the release of a timer job, and the scheduler does not achieve periodic job releases by design. 
However, this does not mean that analytical results are not applicable by default. 
If the following criteria are met, then the scheduler mimics the typical priority-based scheduler, in the sense that it generates the same schedule (although the job releases are not periodic):
\begin{itemize}
  \item[C1] Whenever the \emph{timestamp} of a timer is reached, this eventually leads to a job release, and the \emph{timestamp} is increased by exactly one period. (That is, no job release is skipped.)
  \item[C2] After a \emph{timestamp} is reached, then a job must be released before the next scheduling decision is made (where it will be considered eligible). (That is, jobs are released fast enough.)
\end{itemize}
If these conditions are satisfied, the schedule coincides with the schedule obtained by a typical priority-based scheduler.

Under our scheduler with the \textbf{RE} option, whenever a job finishes, the tasks in the \emph{timer release queue} are reordered, and released jobs are added to the \emph{timer ready queue} before a scheduling decision is made. 
That is, C2 is naturally fulfilled. 
We achieve criteria C1 if a job execution does not span over two possible timestamps.
That is, if we assume that 
\begin{equation}
  C_i < T_j 
\end{equation}
for all $\tau_j \neq \tau_i$, 
then both criteria C1 and C2 are fulfilled, and our scheduler indeed generates schedules that coincide with the schedules obtained by typical priority-based schedulers.
Since the scheduling decision is made only after completing job execution and not during job execution, this behaves like a non-preemptive scheduling mechanism.

\subsection{Bounding the Releaser Overhead}
\label{sec:compatibility:overhead}

In this subsection, we provide bounds on the amount of overhead that the releaser can introduce into the system. 
This overhead has to be accounted for when applying the results in the literature for non-preemptive schedulers.

We assume that releasing a job takes up to $\delta > 0$ time units.
Then, if a job is released into the \emph{events queue} or \emph{timer ready queue}, this prolongs the execution time of the job that is currently running or that is about to start right after the release.
Hence, by counting the maximal number of released jobs right before and during the execution of a job, we obtain a bound on the overhead.
We denote by $\Delta_i$ the overhead that jobs of $\tau_i$ can experience.

With the \textbf{RE} option, releasing jobs cannot happen during the execution of a job, but may only prolong the execution right before. 
Only one job of each task can be released into the \emph{timer ready queue}, leading to an overhead of 
\begin{equation}
  \Delta_i \leq n\cdot \delta
\end{equation}
time units, where $n$ is the number of tasks in the system.
Please note that this covers the time for the release of the job that is about to start.

With the \textbf{RO} option, job releases can be done whenever a timestamp is reached.
That is, releasing can also take place during job execution. 
The number of timestamps that can occur for a task $\tau_j$ during an interval of length $t$ is $\ceiling{\frac{t}{T_j}}$.
Let $t_0 \in \Rbb_{>0}$ be the lowest positive real number such that $t_0 \geq C_i + \sum_{j=1}^n \ceiling{\frac{t_0}{T_j}} \cdot \delta$.
Then $t_0$ is the amount of time that a job of $\tau_i$ can execute, including the overhead from releasing other tasks.
Hence, the release overhead for task $\tau_i$ is bounded by
\begin{equation}\label{eq:overhead_bound_RO:first}
  \Delta_i \leq \sum_{j=1}^n \ceiling{\frac{t_0}{T_j}} \cdot \delta.
\end{equation}

This bound on the overhead $\Delta_i$ can be tightened further if the following two conditions hold:
\begin{itemize}
  \item The task set has constrained deadlines.
  \item The task set is already shown to be schedulable (e.g., by using the bound on the overhead from Equation~\eqref{eq:overhead_bound_RO:first} together with a schedulability test from Section~\ref{sec:compatibility:wcrt}).
\end{itemize}
Under these conditions, every other task can only release one job during the execution of a job of $\tau_i$.
Otherwise, assuming a task $\tau_j$ releases two jobs at timestamps $t_1$ and $t_2= t_1+T_J$ during the execution of a job of $\tau_i$, then the job of $\tau_j$ can not be executed until time $t_2$.
Hence, it would miss its constrained deadline, which violates the schedulability of the task set.
Therefore, under the two conditions, the overhead for the releaser is upper bounded by 
\begin{equation}
  \Delta_i \leq n \cdot \delta,
\end{equation}
where $n$ is the number of tasks in the system.

We note that it is unsafe to apply the tightened bound directly for calculating a response time bound and checking the schedulability based on that.
However, when the response time is already shown to be $\leq D_i$, we can use this formula to tighten the response time bound further.
This can be beneficial when applying analyses that utilize the worst-case response time (cf. Section~\ref{sec:compatibility:e2e}).

\subsection{Worst-Case Response Time Analysis}
\label{sec:compatibility:wcrt}

In the following, we analyze the WCRT of timer tasks under our scheduler. 
We assume that the overhead is accounted for by prolonging the WCET, i.e., by redefining $C_i := C_i + \Delta_i$.
Following our discussion in Section~\ref{sec:compatibility:behavior}, the timer tasks $\tau_i \in \Tbb$ behave like non-preemptive periodic tasks with period $T_i$.
Therefore, literature results for non-preemptive periodic tasks (cf.~\cite{DBLP:journals/rts/DavisBBL07,DBLP:conf/rtss/NasriB17,george1996EDFNP,DBLP:conf/rtss/JeffaySM91,DBLP:journals/rts/TindellBW94}) are applicable.

Although this pertains to any scheduling algorithm that can be modeled by reordering the \emph{events queue} or \emph{timer ready queue}, we focus specifically on Fixed-Priority (FP) and Earliest-Deadline-First (EDF) scheduling.
Although we present only a few results that we utilized in the evaluation, the results of this work are not limited to only these.

The following theorem is a reformulation of the schedulability test for non-preemptive FP scheduling presented for example in the work by von der Br\"uggen et al. \cite{DBLP:conf/ecrts/BruggenCH15}. 

\begin{theorem}[Non-preemptive FP, Reformulated from Eq.~(6)~in~\cite{DBLP:conf/ecrts/BruggenCH15}]
  \label{thm:np-fp}
  Assume that the tasks $\Tbb = \{\tau_1, \dots, \tau_n\}$ are ordered by their priority, i.e., $\tau_1$ has the highest priority and $\tau_n$ has the lowest priority.
  If there exists a $t\geq 0 \in \Rbb$ such that $t \leq D_k$ and 
  \begin{equation}\label{eq:np-fp}
    t \geq C_k + \max_{i>k} C_i + \sum_{i<k} \ceiling{\frac{t}{T_i}} \cdot C_i,
  \end{equation}

  then the WCRT $R_k$ of $\tau_k \in \Tbb$ under non-preemptive FP scheduling is upper bounded by $t$.
\end{theorem}

That is, an upper bound on the WCRT can be computed using fixed-point iteration, starting with $t=0$, to find the smallest $t$ that fulfills Equation~\eqref{eq:np-fp}.
Moreover, if we can find a response time upper bound $R_k \leq D_k$ for all $\tau_k \in \Tbb$, then we know that the task set is schedulable.

Bounds under non-preemptive EDF have for example been explored by George et al.~\cite{george1996EDFNP}.
In the following, we restate their theorem for completeness. 
Please note that while their work considers the discrete time domain, we formulate it for the continuous time domain. 

\begin{theorem}[Non-preemptive EDF, Reformulated from~\cite{george1996EDFNP}]
  \label{thm:np-edf}
  The task set is schedulable under non-preemptive EDF if for all $t>0$ the equation
  \begin{equation}\label{eq:np-edf}
    \max_{\tau_i, D_i>t} C_i + \sum_{\tau_i \in \Tbb} \mathit{dbf}_i(t) \leq t
  \end{equation}
  holds, where $\mathit{dbf}_i$ is the \emph{demand bound function} of task $\tau_i$, defined by 
  $\mathit{dbf}_i(t):= \max\left(0, \floor{\frac{t-D_i}{T_i}+1}\right) \cdot C_i$.
\end{theorem}

We note that when a task set is shown to be schedulable under non-preemptive EDF, then we can also claim an upper bound on the WCRT, namely $R_i \leq D_i$.

\subsection{Application of End-to-End Latency Analysis}
\label{sec:compatibility:e2e}

One of the predominantly studied metrics for ROS~2 applications is end-to-end latency.
That is, given a sequence of tasks $E = (\tau_{i_1} \to \dots \tau_{i_N})$, a so-called \emph{cause-effect chain}, then we are interested in the amount of time that data needs to traverse the cause-effect chain. 
Specifically, the Maximum Reaction Time (MRT), i.e., the amount of time until an external cause is fully processed, and the Maximum Data Age (MDA), i.e., the age of data utilized in an actuation, have been analyzed extensively in the literature.
While there are only a few analytical results for the standard ROS~2 application~\cite{DBLP:conf/rtss/TeperGUBC22,teper2024rtas}
there is a large body of literature results for typical periodic tasks~\cite{DBLP:conf/dac/DavareZNPKS07,DBLP:conf/rtcsa/BeckerDMBN16,DBLP:journals/jsa/BeckerDMBN17,DBLP:conf/ecrts/HamannD0PW17,DBLP:conf/etfa/KlodaBS18,duerrCASES,DBLP:conf/rtas/GunzelCUBDC21,DBLP:conf/rtns/GohariNV22,DBLP:conf/dac/BiLRWLT22,DBLP:conf/rtns/GunzelUCC23,DBLP:conf/ecrts/GunzelTCBC23,DBLP:journals/tecs/GunzelCUBDC23}. 
Our scheduler enables such results and insights from end-to-end analysis for periodic task systems.
Recently, it has been shown that MRT and MDA are equivalent. 
Hence, we use $\mathit{Lat}$ to denote the end-to-end latency in general.

One typical approach to derive end-to-end latency in periodic priority-based scheduling (cf.~\cite{DBLP:conf/dac/DavareZNPKS07}\footnote{Although Davare et al.~\cite{DBLP:conf/dac/DavareZNPKS07} formulate this bound using preemptive FP scheduling, this result is universally applicable when utilizing a correct bound on the WCRT.}) is to sum up the task periods and bounds on the WCRT.
The end-to-end latency is upper bounded by:
\begin{equation}
  \label{eq:latency_bound}
  \mathit{Lat} \leq \sum_{j=1}^N \left( T_{i_j} + R_{i_j} \right)
\end{equation}
More specifically, given a cause-effect chain $E$ on the task set $\Tbb$ scheduled by our scheduler, then the following steps result in a bound on the end-to-end latency:
\begin{enumerate}
  \item Calculating the overhead $\Delta_i$ from Section~\ref{sec:compatibility:overhead} and incorporating it in the WCET by $C_i := C_i + \Delta_i$.
  \item Providing an upper bound on the WCRT $R_i$ using results from Section~\ref{sec:compatibility:wcrt}.
  \item Calculating a bound for the end-to-end latency using Equation~\eqref{eq:latency_bound}.
\end{enumerate}

In Section~\ref{sec:experiments}, we compare this bound on the end-to-end latency with the latency that can be analytically guaranteed for the typical ROS~2 scheduler~\cite{DBLP:conf/rtss/TeperGUBC22}.

\section{Subscription Tasks}
\label{sec:subscription}

The previous sections focused on the behavior of timer tasks. 
However, besides timers, ROS~2 allows tasks to be triggered through the built-in Data Distribution Service (DDS).
That is, when a job finishes its execution it can publish a message to a certain predefined topic, and all subscription tasks subscribing to this topic are activated, such that they can be released.
Since subscription tasks are widely used in practice, this section explains how our findings can be extended to systems involving subscription tasks.

In the default ROS~2 executor, all activated subscriptions are released at the polling point.
In the events executor, subscriptions are only processed by the \emph{default thread}, and before every scheduling decision of the \emph{default thread}, it releases all activated subscriptions by moving them to the \emph{events queue}.
To that end, we first consider the \textbf{RE} option, and afterward, we consider the \textbf{RO} option.
This section does not aim to provide specific analytical results, but rather links different scenarios to the corresponding real-time systems literature.

\textbf{RE Option}: 
For this configuration, timers are handled completely separately by the \emph{timer management thread}, while subscriptions are fully handled by the \emph{default thread}.
  
As subscriptions are not activated or released periodically, the
analysis of this work is not directly applicable.  If a minimum
inter-arrival time $>0$, i.e., a minimum time between two publications
of a topic that the subscription subscribes to can be determined, then
the literature on sporadic non-preemptive tasks may be applicable.
However, to apply the literature it has to be ensured that every
activation also leads to a job release, i.e., no jobs are lost.
This is beyond the scope of this paper.

\textbf{RO Option}:
In this configuration, both timers and subscriptions are released to the prioritized \emph{events queue} of the \emph{default thread}.
For this option, we use our proposal from Section~\ref{sec:release-only} to give the \emph{timer management thread} a higher priority than the \emph{default thread}, to ensure that jobs are released before scheduling decisions are made.
Since the execution of all jobs is handled by the same thread, the literature on non-preemptive scheduling is applicable.
However, to apply the results for the timer tasks, the subscription tasks have to be taken into account as well.
To that end, a minimum inter-arrival time $>0$ has to be determined for each subscription task. 
To ensure that no job is lost, the minimum inter-arrival times and task periods have to be larger than the WCET of any task.

One special case that we would like to emphasize is when tasks are arranged in \emph{sequences}.
That is, each task invokes the activation of at most one subscription task, and each subscription task can be activated by exactly one task.
An example of such sequences can look as follows:
\begin{itemize}
  \item $\text{sequence}_1 = \text{timer}_1 \to \text{subscription}_1 \to \text{subscription}_2$
  \item $\text{sequence}_2 = \text{timer}_2 \to \text{subscription}_3 \to \text{subscription}_4$
\end{itemize}
We consider the special case where the first task of a sequence is a timer task, as this guarantees a minimum inter-arrival time for the subsequent subscription tasks.
Furthermore, all subsequent tasks are subscription tasks since they subscribe to a topic by definition.
Each job of the timer task of a sequence releases one job of the subsequent subscription tasks.
When considering the whole sequence as a task, with timer tasks and subscription tasks only being subtasks of the sequence, then this relates to the literature on limited preemptive scheduling~\cite{DBLP:journals/tii/ButtazzoBY13}. 

Each subtask is scheduled non-preemptively, and between subtasks, the task can be preempted.
The benefit of considering sequences is that there is no need to determine the minimum inter-arrival time of subscription tasks.
However, as soon as the data dependencies of the DDS become more complex, this approach is not applicable anymore. 

We further note that, in case the DDS introduces a certain delay to release the subsequent subscription\cite{kronauer2021latency}, another lower priority job may be started in between the execution of two non-preemptive subtasks. 
In that case, the additional delay can be modeled as self-suspension~\cite{DBLP:journals/rts/ChenNHYBBLRRARN19}.
More specifically, the set of sequences behaves like a set of segmented self-suspending tasks scheduled non-preemptively on a single core.
This delay can be assumed to be zero if zero-copy middleware is used, or if the DDS communication is synchronous. In these cases, downstream tasks are activated by the time the upstream task finishes and are immediately eligible for execution.

\section{Evaluation}\label{sec:experiments}

In this section, we evaluate our proposed modifications of the ROS~2
events executor and the corresponding analyses. 
The evaluation consists of three contributions:
\begin{itemize}
  \item We confirm \textbf{compatibility with scheduling theory} on non-preemptive schedulers for periodic tasks in Section~\ref{sec:experiments_1}. More specifically, we demonstrate for a timers-only task set that the analytical results of Section~\ref{sec:compatibility} are applicable. 
  That is, the modified ROS~2 events executor does not drop jobs, and it respects the worst-case response time bounds for non-preemptive scheduling presented in Section~\ref{sec:compatibility:wcrt}.
  \item In Section~\ref{sec:experiments_2}, we show that the \textbf{bounds on end-to-end latency} for cause-effect chains, enabled by the theory of non-preemptive scheduling bounds, are tighter than the state-of-the-art analytical end-to-end latency bounds for the default executor. 
  To that end, we apply the analyses to synthetic task sets using the WATERS~\cite{kramer2015real} benchmark.
  \item We compare the performance of our modified executor using the RM scheduling policy with the default executor in the Autoware reference system in Section~\ref{sec:experiments_3}.
  The Autoware reference system includes both timers and subscriptions and demonstrates that we can achieve lower end-to-end latency than the other executor options provided natively by ROS~2.
\end{itemize}
Our experiments feature the following executor configurations:
\begin{itemize}
  \item default executor (\textbf{Default}), 
  \item static default single-threaded Executor (\textbf{Static}), %
  \item unmodified default events executor (\textbf{Events}), %
  \item our rate-monotonic events executor (\textbf{RM~(RO)} and
    \textbf{RM~(RE)}), and %
  \item our earliest-deadline-first events executor (\textbf{EDF~(RO)}
    and \textbf{EDF~(RE)}). %
\end{itemize}
We include the static default executor for completeness. It behaves like the default executor, but with less overhead, as it does not have to handle dynamic system configurations.

\subsection{Compatibility with Scheduling Theory}
\label{sec:experiments_1}

In this subsection, we demonstrate that existing results on non-preemptive scheduling apply to our scheduler, and analytical results correctly predict the behavior of the system.

To that end, we examine a timers-only task set that consists of three types of components, and seven nodes in total. The system has four cameras, two LiDARs, and an IMU. Each node has one timer task, and each component is independent of the others, meaning that they do not pass data between each other and have no precedence constraints. 
By varying the maximum execution time of the tasks, we configure the system to have a utilization of 60\%, 80\%, or 90\%.
Our three configurations are as follows:

\medskip%
\textbf{60\% utilization}
\begin{itemize}
  \item 4 Camera Nodes: 84ms period, 10ms execution each
  \item 2 LiDAR Nodes: 200ms period, 10ms execution each
  \item 1 IMU Node, 30ms period, 1ms execution
\end{itemize}

\textbf{80\% utilization}
\begin{itemize}
  \item 4 Camera Nodes: 84ms period, 14ms execution each
  \item 2 LiDAR Nodes: 200ms period, 10ms execution each
  \item 1 IMU Node, 30ms period, 1ms execution
\end{itemize}

\textbf{90\% utilization}
\begin{itemize}
  \item 4 Camera Nodes: 84ms period, 16ms execution each
  \item 2 LiDAR Nodes: 200ms period, 10ms execution each
  \item 1 IMU Node, 30ms period, 1ms execution
\end{itemize}

Task periods stay consistent across all utilization levels, so all task sets have the same hyperperiod of 4.2 seconds. Each task set is run for 5 minutes, elapsing over 70 hyperperiods.
We repeat this for each executor design.

Each of these tasksets are pinned to a single core of a Raspberry Pi Model 4B with 4GB of RAM running Ubuntu~22.04 and ROS~2 Humble. \ref{sec:experiments_1} uses one core, while \ref{sec:experiments_3} uses two cores with no thread pinning. \ref{sec:experiments_2} has one executor thread running on one processing core.

\begin{figure}
  \centering
  \includegraphics[width=0.48\textwidth]{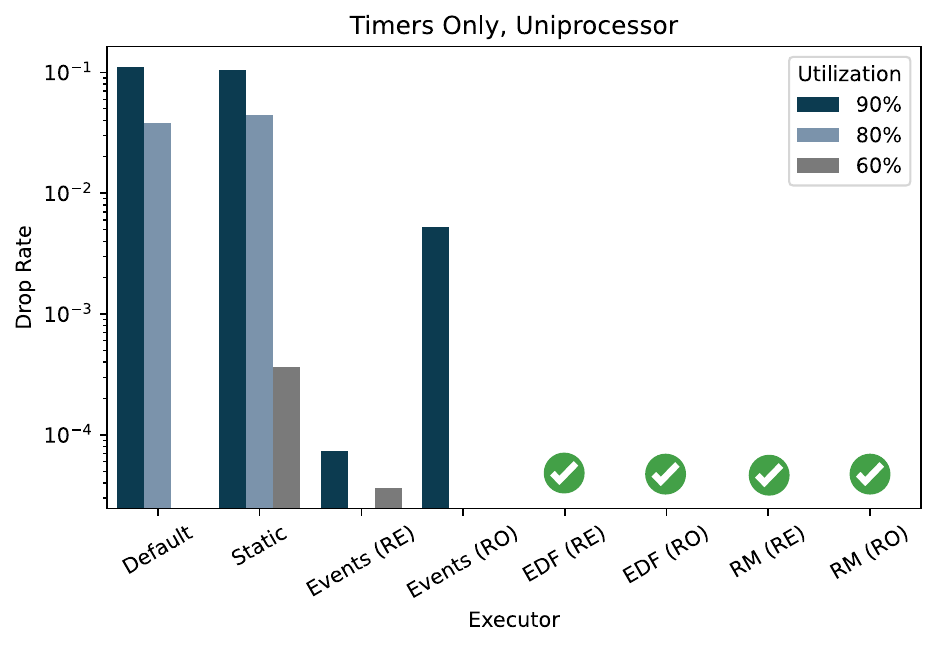}
  \caption{Dropped jobs on timers-only task sets: A checkmark indicates no dropped jobs across all utilization levels.}
  \label{fig:dropped_jobs_timers_only}
\end{figure}

\medskip%
\textbf{Number of Dropped Jobs:}
The rate of dropped jobs for different configurations is shown in Figure~\ref{fig:dropped_jobs_timers_only}.
We can see that the default, static, and unmodified events executor repeatedly drop jobs. 
In some cases, a drop rate of $10\%$ is observed.
Following our analytical results of Section~\ref{sec:compatibility:behavior}, our proposed events executor do not have any job drops for any configuration of the RM and EDF events executors.

\medskip%
\textbf{Response Time Bounds:}
In the following, we analytically derive upper bounds on the worst-case response time, following Section~\ref{sec:compatibility}, and then confirm that the response time bounds are not violated by our RM events executor.

For the analysis, we measured that a job release takes up to $\delta = 0.12ms$.
The total overhead $\Delta_i$ for a timer task $\tau_i$ can be determined by Equation~\eqref{eq:overhead_bound_RO:first}.
Doing the calculations for the task sets of this section, every timer task has an overhead of $\Delta_i = 0.84ms$.
Afterward, we apply Theorem~\ref{thm:np-fp} with the updated worst-case response time bounds $C_i := C_i + \Delta_i$.

We consider implicit-deadline task systems, i.e., the relative deadline of a timer task is equal to its timer period. 
The analyzed worst-case response times for our RM events executor are reported in Table~\ref{table:analytical_wcrt}, indicating no deadline misses.
Thus, we should not observe any job drops, as verified in Figure~\ref{fig:dropped_jobs_timers_only}, and all tasks are guaranteed to meet their deadlines.
\begin{table}
  \centering
  \caption{Analytical worst-case response times (ms) of our RM events executor.}
  \label{table:analytical_wcrt}
  \begin{tabular}{c|rrr|c}
    Utilization & 60\% & 80\% & 90\% & Period\\
    \hline
    IMU & 12.67 & 16.67 & 18.67 & 30\\
    Camera & 57.83 & 75.66 & 83.66 & 84\\
    LiDAR & 70.50 & 149.50 & 167.33 & 200\\
  \end{tabular}
\end{table}

\begin{figure}[t!]
  \centering
  \includegraphics[width=0.48\textwidth]{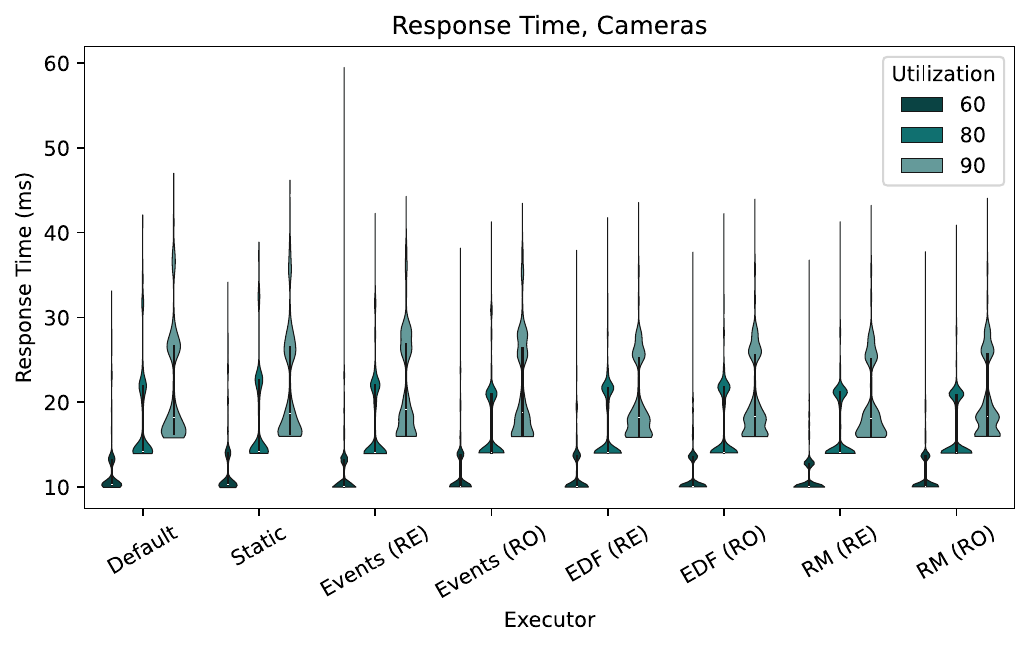}
  \caption{Response Time of Camera Tasks (ms), D=84ms}
  \label{fig:response_time_camera}
\end{figure}

\begin{figure}
  \centering
  \includegraphics[width=0.48\textwidth]{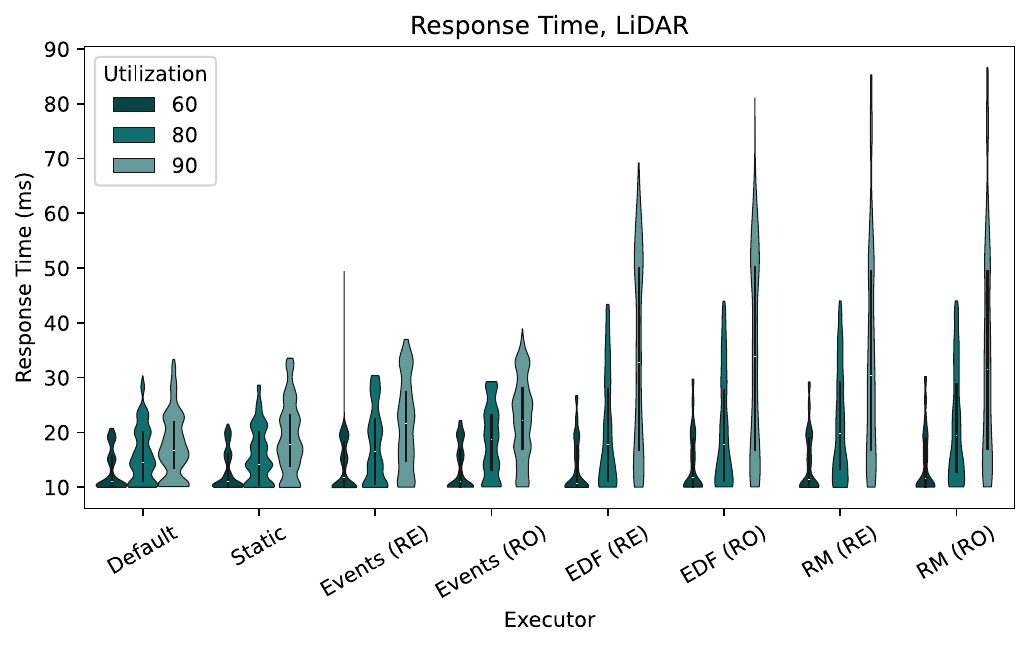}
  \caption{Response Time of LiDAR Tasks (ms), D=200ms}
  \label{fig:response_time_lidar}
\end{figure}

\begin{figure}
  \centering
  \includegraphics[width=0.48\textwidth]{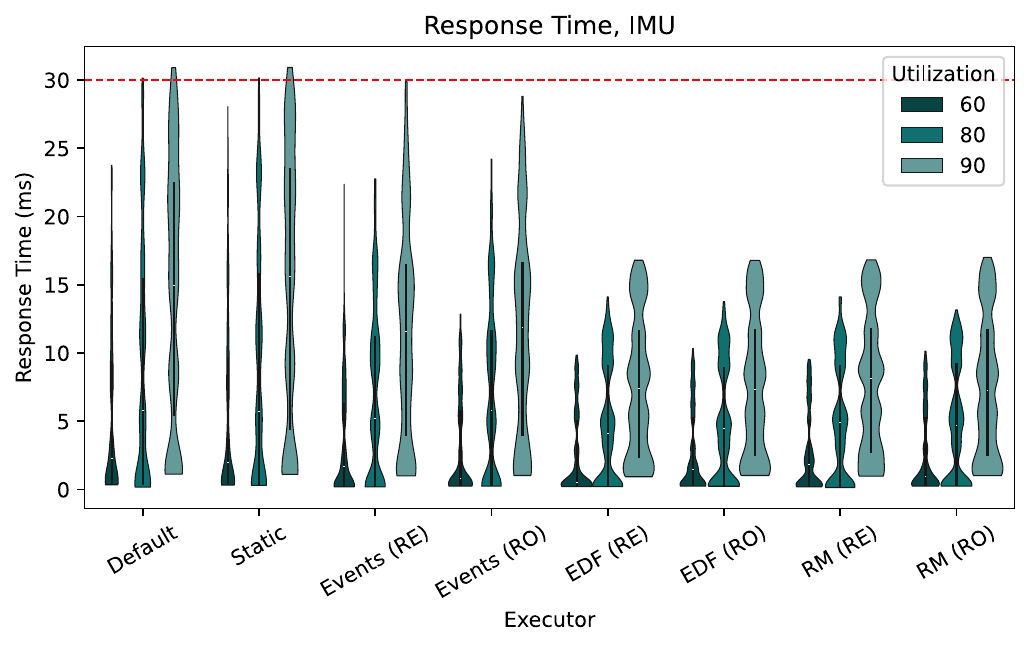}
  \caption{Response Time of IMU Tasks (ms), D=30ms}
  \label{fig:response_time_imu}
\end{figure}

We measured the response times during the experiment.
The aggregated results of the Camera nodes are shown in Fig~\ref{fig:response_time_camera}, the LiDAR nodes in Fig~\ref{fig:response_time_lidar}, and the IMU node in Fig~\ref{fig:response_time_imu}.
As shown in the figures, the response times are consistent with the analytical results, and no deadline overruns are observed.
Furthermore, in Figure~\ref{fig:response_time_imu}, we see that the IMU tasks have deadline overruns on the Default, Static, and Events executors.
We note that Figures~\ref{fig:response_time_camera}, \ref{fig:response_time_lidar} and \ref{fig:response_time_imu} only account for the response times of jobs that are not dropped. 
Hence, for a significant amount of timestamps, the Default, Static, and Events executors do not provide a response within the deadline, as observed in Figure~\ref{fig:dropped_jobs_timers_only}.

In conclusion, our analysis not only correctly indicates the absence of dropped jobs in our RM events executor, but also a correct upper bound on the response time of each task.


\subsection{Tighter End-to-End Bounds for ROS 2}
\label{sec:experiments_2}

In this section, we evaluate the differences between the analytically derived upper bounds of the end-to-end latencies between the default ROS~2 executor and our RM events executor.
The task sets are synthetically generated using the WATERS benchmark~\cite{kramer2015real} for automotive systems. %

We evaluate task sets for utilization levels of 60\%, 80\%, and 90\%.
For each configuration, we generate one thousand task sets, each including 10 to 200 tasks.
To calculate the end-to-end latencies, we generate 5 to 60 chains of tasks, each of which consists of 2 to 15 tasks.

For each task set, we calculate the end-to-end latency for the default ROS~2 executor and our RM (including both RO- and RE-options, as the analysis) events executor, as presented in Section~\ref{sec:compatibility:e2e}. Specifically, we use Equation~\eqref{eq:latency_bound} in Section~\ref{sec:compatibility} to calculate the end-to-end latency of each chain.

The end-to-end latency of the default ROS~2 executor has been analyzed by Teper et al.~\cite{teper2024rtas}.
We use Equation~(15)~of~Lemma~VI.1 and Equation~(17)~of~Lemma~VI.2 of their paper~\cite{teper2024rtas} to get an upper bound on the latency of each timer.
For each chain, we sum up the latency upper bounds of all chain tasks to get the end-to-end latency of the chain.

\begin{figure}
  \centering
  \includegraphics[width=0.48\textwidth]{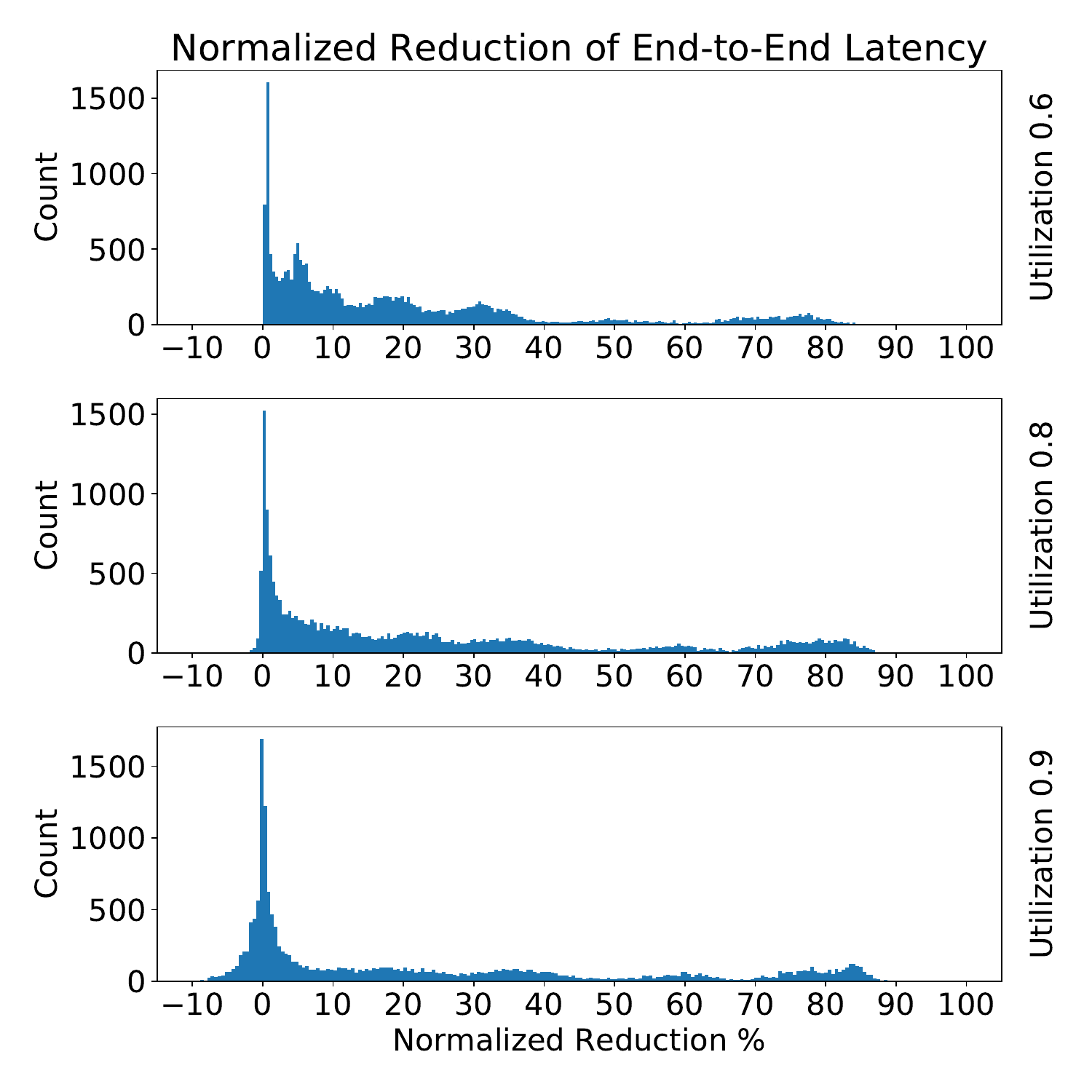}
  \caption{Reduction of the end-to-end latency between the default ROS~2 executor and our RM events executor.}
  \label{fig:latency_comparison}
\end{figure}

For each chain, we calculate the normalized reduction of the end-to-end latency by using $\frac{E_{\mathit{default}} - E_{\mathit{RM}}}{E_{\mathit{default}}}$, where $E_{\mathit{default}}$ ($E_{\mathit{RM}}$, respectively) is the end-to-end latency of the chain under the default executor (under our RM events executor, respectively). The results of the evaluation are shown in Figure~\ref{fig:latency_comparison}, where the $x$-axis is the normalized reduction and the $y$-axis is the number of chains within the binned normalized reduction. 

We can see that the end-to-end latencies of the chains are usually significantly lower when using our RM events executor.
As the utilization level increases, the number of chains with a higher end-to-end latency increases as well. This is because
the default ROS 2 executor is comparatively fair to all callbacks regarding their priorities. When using static-priority scheduling, some callbacks may be assigned lower priorities, leading to an increase of response time (see Figure~\ref{fig:response_time_lidar}), resulting in a negative reduction of end-to-end latencies for chains that include such tasks.
However, the majority of chains have lower end-to-end latency when using the RM events executor, compared to the default ROS~2 executor.
In some cases, we observe a normalized reduction of the end-to-end latency bound of almost $90\%$. Our experiment shows that our design can be applied to different kinds of task sets and potentially leads to latency improvements.

\subsection{Performance of Autoware Reference System}
\label{sec:experiments_3}

In this section, we conduct an end-to-end latency evaluation for the Autoware reference system, detailed in~\cite{referencesystem}, which includes interconnected timers and subscriptions.\footnote{\url{https://github.com/ros-realtime/reference-system/blob/main/autoware_reference_system/README.md}}
The Autoware reference benchmark runs a simulated version of the Autoware application on a Raspberry Pi Model 4B running Ubuntu~22.04 and ROS~2 Humble.
We run the ROS~2 Autoware benchmark on two dedicated cores, whilst the other two cores are handling the remaining services on the operating system.
Specifically, we measure the end-to-end latency of the \textbf{hot path} defined in the benchmark from the front and rear LiDAR sensors to the object collision estimator (c.f. the Autoware reference system), as this latency is a key metric for the responsiveness to crash avoidance.
The data at the start of the hot path is generated at a frequency of 10Hz.

\begin{figure}[t]
    \centering
    \includegraphics[width=0.48\textwidth]{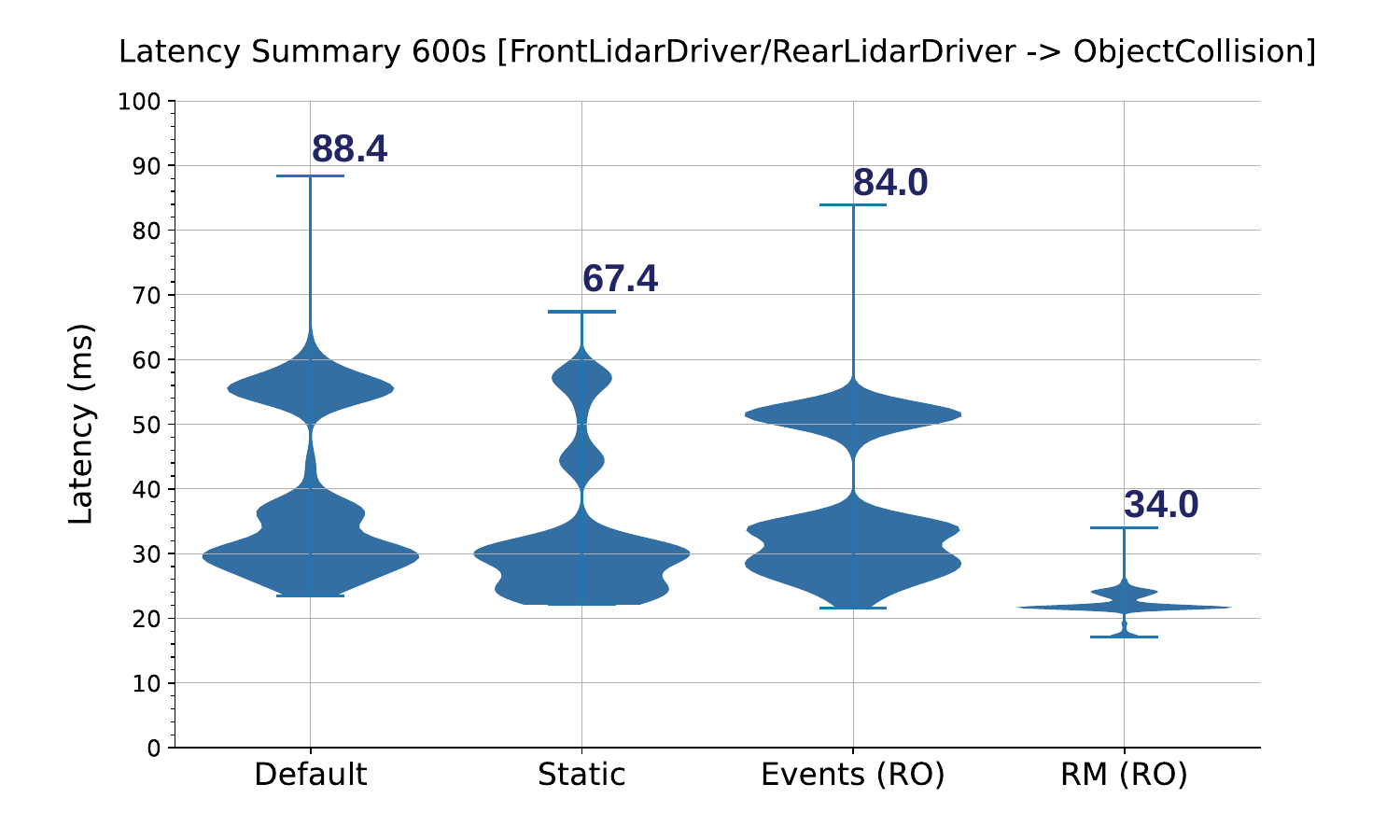}
    \caption{End-to-end latency of the hot path in the Autoware reference system under different executors.}
    \label{fig:executor_latency}
\end{figure}

For our RM events executor, the subscription tasks inherit the highest priority from their corresponding upstream publishers.
Non-preemptive EDF in this scenario is not well specified, as subscription tasks do not have any assigned deadlines, and hence is not evaluated.
For the default events executor and the RM events executor, we use \textbf{RO}, so that all jobs are scheduled by and executed on the \emph{default thread}.

Figure~\ref{fig:executor_latency} shows the violin plot for the latency of the hot path in the Autoware reference system from a test run of $600$ seconds.
Of the executors measured, our RM events executor has a better worst-case latency, significantly outperforming the other executors.
Compared to our work, the default ROS~2 executor has a 2.6x slowdown along the hot path.
Our evaluation shows that our executor design using existing scheduling policies from classical real-time systems can be applied to practical ROS~2 applications, and can provide major benefits in terms of end-to-end latency.

\section{Conclusion}\label{sec:conclusion}

This paper provides deep investigations of the recently introduced
events executor (in 2023) in ROS~2 to provide compatibility with the
classical real-time scheduling theory of periodic task
systems. Specifically, our solution is easy to integrate into existing
ROS~2 systems, since it requires only minor backend modifications of
the events executor, already natively included in ROS~2. Our study
enables the rich literature of the real-time scheduling theory for
non-preemptive schedules to apply to ROS~2. Furthermore, we
intensively validate the feasibility of our modifications for
non-preemptive RM and non-preemptive EDF in the ROS~2 events executor
for several scenarios. The evaluation results show that our ROS~2
events executor with minor modifications can have significant
improvement in terms of dropped jobs, worst-case response time,
end-to-end latency, and performance, in comparison to the default
ROS~2 executor.

We note that the emphasis of our paper is the compatibility of the ROS~2
events executor with the classical scheduling theory of periodic
tasks. As a result, we focus on the transformation of a timer into a
periodic task properly and only sketch how the event-triggered
subscription tasks can be integrated. The interaction of the built-in
DDS and the events executor when both timers and subscriptions exist
requires further investigation. Furthermore, although we have
successfully implemented non-preemptive EDF into the events
executor, extending non-preemptive EDF of timers to deal with
subscription tasks is not well specified, as subscription tasks do not
have any assigned deadlines. Further investigations on the
combinations of non-preemptive EDF for timers and other scheduling
strategies for subscription tasks are needed.

\bibliographystyle{plain}
\bibliography{references}

\begin{thebibliography}{10}

\bibitem{arafat2022}
Abdullah~Al Arafat, Sudharsan Vaidhun, Kurt~M. Wilson, Jinghao Sun, and Zhishan
  Guo.
\newblock {Response time analysis for dynamic priority scheduling in ROS2}.
\newblock In {\em Proceedings of the 59th ACM/IEEE Design Automation
  Conference}, page 301–306, 2022.

\bibitem{DBLP:conf/rtcsa/BeckerDMBN16}
Matthias Becker, Dakshina Dasari, Saad Mubeen, Moris Behnam, and Thomas Nolte.
\newblock Synthesizing job-level dependencies for automotive multi-rate effect
  chains.
\newblock In {\em International Conference on Embedded and Real-Time Computing
  Systems and Applications (RTCSA)}, pages 159--169, 2016.

\bibitem{DBLP:journals/jsa/BeckerDMBN17}
Matthias Becker, Dakshina Dasari, Saad Mubeen, Moris Behnam, and Thomas Nolte.
\newblock End-to-end timing analysis of cause-effect chains in automotive
  embedded systems.
\newblock {\em Journal of Systems Architecture - Embedded Systems Design},
  80:104--113, 2017.

\bibitem{DBLP:conf/dac/BiLRWLT22}
Ran Bi, Xinbin Liu, Jiankang Ren, Pengfei Wang, Huawei Lv, and Guozhen Tan.
\newblock Efficient maximum data age analysis for cause-effect chains in
  automotive systems.
\newblock In {\em {DAC}}, pages 1243--1248. {ACM}, 2022.

\bibitem{blass2021}
Tobias Blass, Daniel Casini, Sergey Bozhko, and Bj{\"o}rn~B. Brandenburg.
\newblock {A ROS 2 Response-Time Analysis Exploiting Starvation Freedom and
  Execution-Time Variance}.
\newblock In {\em Proceedings of the 42nd {{Real}}-Time {{Systems Symposium}}
  ({{RTSS}})}, 2021.

\bibitem{DBLP:journals/tii/ButtazzoBY13}
Giorgio~C. Buttazzo, Marko Bertogna, and Gang Yao.
\newblock Limited preemptive scheduling for real-time systems. {A} survey.
\newblock {\em {IEEE} Trans. Ind. Informatics}, 9(1):3--15, 2013.

\bibitem{casini2019response}
Daniel Casini, Tobias Bla{\ss}, Ingo L{\"u}tkebohle, and Bj{\"o}rn Brandenburg.
\newblock Response-time analysis of ros 2 processing chains under
  reservation-based scheduling.
\newblock In {\em 31st Euromicro Conference on Real-Time Systems}, pages 1--23.
  Schloss Dagstuhl, 2019.

\bibitem{DBLP:journals/rts/ChenNHYBBLRRARN19}
Jian{-}Jia Chen, Geoffrey Nelissen, Wen{-}Hung Huang, Maolin Yang,
  Bj{\"{o}}rn~B. Brandenburg, Konstantinos Bletsas, Cong Liu, Pascal Richard,
  Fr{\'{e}}d{\'{e}}ric Ridouard, Neil~C. Audsley, Raj Rajkumar, Dionisio
  de~Niz, and Georg von~der Br{\"{u}}ggen.
\newblock Many suspensions, many problems: a review of self-suspending tasks in
  real-time systems.
\newblock {\em Real Time Syst.}, 55(1):144--207, 2019.

\bibitem{choi2021picas}
Hyunjong Choi, Yecheng Xiang, and Hyoseung Kim.
\newblock Picas: New design of priority-driven chain-aware scheduling for ros2.
\newblock In {\em 2021 IEEE 27th Real-Time and Embedded Technology and
  Applications Symposium (RTAS)}, pages 251--263. IEEE, 2021.

\bibitem{DBLP:conf/dac/DavareZNPKS07}
Abhijit Davare, Qi~Zhu, Marco~Di Natale, Claudio Pinello, Sri Kanajan, and
  Alberto~L. Sangiovanni{-}Vincentelli.
\newblock Period optimization for hard real-time distributed automotive
  systems.
\newblock In {\em Design Automation Conference, {DAC}}, pages 278--283, 2007.

\bibitem{DBLP:journals/rts/DavisBBL07}
Robert~I. Davis, Alan Burns, Reinder~J. Bril, and Johan~J. Lukkien.
\newblock Controller area network {(CAN)} schedulability analysis: Refuted,
  revisited and revised.
\newblock {\em Real Time Syst.}, 35(3):239--272, 2007.

\bibitem{DBLP:journals/rts/DavisTGDPC18}
Robert~I. Davis, Abhilash Thekkilakattil, Oliver Gettings, Radu Dobrin,
  Sasikumar Punnekkat, and Jian-Jia Chen.
\newblock Exact speedup factors and sub-optimality for non-preemptive
  scheduling.
\newblock {\em Real Time Syst.}, 54(1):208--246, 2018.

\bibitem{duerrCASES}
Marco D{\"{u}}rr, Georg von~der Br{\"{u}}ggen, Kuan-Hsun Chen, and Jian-Jia
  Chen.
\newblock End-to-end timing analysis of sporadic cause-effect chains in
  distributed systems.
\newblock {\em {ACM} Trans. Embedded Comput. Syst. (Special Issue for CASES)},
  18(5s):58:1--58:24, 2019.

\bibitem{george1996EDFNP}
Laurent George, Nicolas Rivierre, and Marco Spuri.
\newblock Preemptive and non-preemptive real-time uni-processor scheduling.
\newblock {\em Research Report {RR-2966}, Project {REFLECS}}, 1996.

\bibitem{DBLP:conf/rtns/GohariNV22}
Pourya Gohari, Mitra Nasri, and Jeroen Voeten.
\newblock Data-age analysis for multi-rate task chains under timing
  uncertainty.
\newblock In {\em {RTNS}}, pages 24--35. {ACM}, 2022.

\bibitem{DBLP:conf/rtas/GunzelCUBDC21}
Mario G{\"{u}}nzel, Kuan{-}Hsun Chen, Niklas Ueter, Georg von~der
  Br{\"{u}}ggen, Marco D{\"{u}}rr, and Jian{-}Jia Chen.
\newblock Timing analysis of asynchronized distributed cause-effect chains.
\newblock In {\em {RTAS}}, pages 40--52. {IEEE}, 2021.

\bibitem{DBLP:journals/tecs/GunzelCUBDC23}
Mario G{\"{u}}nzel, Kuan{-}Hsun Chen, Niklas Ueter, Georg von~der
  Br{\"{u}}ggen, Marco D{\"{u}}rr, and Jian{-}Jia Chen.
\newblock Compositional timing analysis of asynchronized distributed
  cause-effect chains.
\newblock {\em {ACM} Trans. Embed. Comput. Syst.}, 22(4):63:1--63:34, 2023.

\bibitem{DBLP:conf/ecrts/GunzelTCBC23}
Mario G{\"{u}}nzel, Harun Teper, Kuan{-}Hsun Chen, Georg von~der Br{\"{u}}ggen,
  and Jian{-}Jia Chen.
\newblock On the equivalence of maximum reaction time and maximum data age for
  cause-effect chains.
\newblock In {\em {ECRTS}}, volume 262 of {\em LIPIcs}, pages 10:1--10:22.
  Schloss Dagstuhl - Leibniz-Zentrum f{\"{u}}r Informatik, 2023.

\bibitem{DBLP:conf/rtns/GunzelUCC23}
Mario G{\"{u}}nzel, Niklas Ueter, Kuan{-}Hsun Chen, and Jian{-}Jia Chen.
\newblock Timing analysis of cause-effect chains with heterogeneous
  communication mechanisms.
\newblock In {\em {RTNS}}, pages 224--234. {ACM}, 2023.

\bibitem{DBLP:conf/ecrts/HamannD0PW17}
Arne Hamann, Dakshina Dasari, Simon Kramer, Michael Pressler, and Falk Wurst.
\newblock Communication centric design in complex automotive embedded systems.
\newblock In {\em Euromicro Conference on Real-Time Systems, {ECRTS}}, pages
  10:1--10:20, 2017.

\bibitem{eventexecutor}
iRobot.
\newblock Events executor.
\newblock \url{https://github.com/irobot-ros/events-executor}, 2022.

\bibitem{DBLP:conf/rtss/JeffaySM91}
Kevin Jeffay, Donald~F. Stanat, and Charles~U. Martel.
\newblock On non-preemptive scheduling of period and sporadic tasks.
\newblock In {\em {RTSS}}, pages 129--139. {IEEE} Computer Society, 1991.

\bibitem{jiang2022}
Xu~Jiang, Dong Ji, Nan Guan, Ruoxiang Li, Yue Tang, and Yi~Wang.
\newblock {Real-Time Scheduling and Analysis of Processing Chains on
  Multi-threaded Executor in ROS 2}.
\newblock In {\em 2022 IEEE Real-Time Systems Symposium (RTSS)}, pages 27--39,
  2022.

\bibitem{DBLP:conf/etfa/KlodaBS18}
Tomasz Kloda, Antoine Bertout, and Yves Sorel.
\newblock Latency analysis for data chains of real-time periodic tasks.
\newblock In {\em {IEEE} International Conference on Emerging Technologies and
  Factory Automation, {ETFA}}, pages 360--367, 2018.

\bibitem{kramer2015real}
Simon Kramer, Dirk Ziegenbein, and Arne Hamann.
\newblock Real world automotive benchmarks for free.
\newblock In {\em International Workshop on Analysis Tools and Methodologies
  for Embedded and Real-time Systems (WATERS)}, 2015.

\bibitem{kronauer2021latency}
Tobias Kronauer, Joshwa Pohlmann, Maximilian Matth{\'e}, Till Smejkal, and
  Gerhard Fettweis.
\newblock Latency analysis of ros2 multi-node systems.
\newblock In {\em 2021 IEEE international conference on multisensor fusion and
  integration for intelligent systems (MFI)}, pages 1--7. IEEE, 2021.

\bibitem{lange2018cbgexecutor}
Ralph Lange.
\newblock Mixed real-time criticality with ros2 - the callback-group-level
  executor.
\newblock ROSCon Lightning Talk, 2018.

\bibitem{liu73scheduling}
C.~L. Liu and James~W. Layland.
\newblock Scheduling algorithms for multiprogramming in a hard-real-time
  environment.
\newblock {\em Journal of the ACM}, 20(1):46--61, 1973.

\bibitem{DBLP:conf/rtss/NasriB17}
Mitra Nasri and Bj{\"{o}}rn~B. Brandenburg.
\newblock An exact and sustainable analysis of non-preemptive scheduling.
\newblock In {\em {RTSS}}, pages 12--23. {IEEE} Computer Society, 2017.

\bibitem{sobhani2023}
Hoora Sobhani, Hyunjong Choi, and Hyoseung Kim.
\newblock {Timing Analysis and Priority-driven Enhancements of ROS 2
  Multi-threaded Executors}.
\newblock In {\em 2023 IEEE 29th Real-Time and Embedded Technology and
  Applications Symposium (RTAS)}, pages 106--118, 2023.

\bibitem{tang2020response}
Yue Tang, Zhiwei Feng, Nan Guan, Xu~Jiang, Mingsong Lv, Qingxu Deng, and Wang
  Yi.
\newblock Response time analysis and priority assignment of processing chains
  on ros2 executors.
\newblock In {\em 2020 IEEE Real-Time Systems Symposium (RTSS)}, pages
  231--243. IEEE, 2020.

\bibitem{teper2024rtas}
Harun Teper, Tobias Betz, Mario Günzel, Dominic Ebner, Georg von~der Brüggen,
  Johannes Betz, and Jian-Jia Chen.
\newblock End-to-end timing analysis and optimization of multi-executor {ROS} 2
  systems.
\newblock In {\em {RTAS}}, 2024.

\bibitem{teper2023}
Harun Teper, Tobias Betz, Georg von~der Brüggen, Kuan-Hsun Chen, Johannes
  Betz, and Jian-Jia Chen.
\newblock Timing-aware ros 2 architecture and system optimization.
\newblock In {\em 2023 IEEE 29th International Conference on Embedded and
  Real-Time Computing Systems and Applications (RTCSA)}, pages 206--215, 2023.

\bibitem{DBLP:conf/rtss/TeperGUBC22}
Harun Teper, Mario G{\"{u}}nzel, Niklas Ueter, Georg von~der Br{\"{u}}ggen, and
  Jian{-}Jia Chen.
\newblock End-to-end timing analysis in {ROS2}.
\newblock In {\em {RTSS}}, pages 53--65. {IEEE}, 2022.

\bibitem{teper2022}
Harun Teper, Mario Günzel, Niklas Ueter, Georg von~der Brüggen, and Jian-Jia
  Chen.
\newblock {End-To-End Timing Analysis in ROS2}.
\newblock In {\em 2022 IEEE Real-Time Systems Symposium (RTSS)}, pages 53--65,
  2022.

\bibitem{referencesystem}
ROS2~Real time Working~Group.
\newblock Autoware reference system.
\newblock \url{https://github.com/ros-realtime/reference-system}, 2022.

\bibitem{DBLP:journals/rts/TindellBW94}
Ken Tindell, Alan Burns, and Andy~J. Wellings.
\newblock An extendible approach for analyzing fixed priority hard real-time
  tasks.
\newblock {\em Real Time Syst.}, 6(2):133--151, 1994.

\bibitem{DBLP:conf/ecrts/BruggenCH15}
Georg von~der Br{\"{u}}ggen, Jian-Jia Chen, and Wen-Hung Huang.
\newblock Schedulability and optimization analysis for non-preemptive static
  priority scheduling based on task utilization and blocking factors.
\newblock In {\em 27th Euromicro Conference on Real-Time Systems, {ECRTS} 2015,
  Lund, Sweden, July 8-10, 2015}, pages 90--101, 2015.

\end{thebibliography}

\end{document}